\documentclass[10pt,twocolumn,letterpaper]{article}

\usepackage[]{cvpr}      
\usepackage{multirow}
\usepackage{footmisc}
\usepackage{float}

\usepackage{color}
\usepackage[accsupp]{axessibility} 

\usepackage{marvosym} 
\usepackage{lipsum} 
\newcommand\blfootnote[1]{%
  \begingroup
  \renewcommand\thefootnote{}\footnote{#1}%
  \addtocounter{footnote}{-1}%
  \endgroup
}

\usepackage{xcolor}

\definecolor{cvprblue}{rgb}{0.21,0.49,0.74}
\usepackage[pagebackref,breaklinks,colorlinks,allcolors=cvprblue]{hyperref}

\def\paperID{4132} 
\def\confName{CVPR}
\def\confYear{2025}

\title{Linguistics-aware Masked Image Modeling for Self-supervised Scene Text Recognition}

\author{Yifei Zhang\textsuperscript{1,3}, Chang Liu\textsuperscript{4}, Jin Wei\textsuperscript{5}, Xiaomeng Yang\textsuperscript{6}, Yu Zhou\textsuperscript{2 \Letter}, Can Ma\textsuperscript{1}, Xiangyang Ji\textsuperscript{4} \\
\textsuperscript{1}{Institute of Information Engineering, Chinese Academy of Sciences} \\
\textsuperscript{2}VCIP \& TMCC \& DISSec, College of Computer Science, Nankai University \\
\textsuperscript{3}School of Cyber Security, University of Chinese Academy of Sciences \\
\textsuperscript{4}Department of Automation and BNRist, Tsinghua University \\
\textsuperscript{5}Lenovo Research 
\textsuperscript{6}College of Engineering, Northeastern University \\
{\tt\small zhangyifei0115@iie.ac.cn, yzhou@nankai.edu.cn}
}

\begin{document}

\twocolumn[{%
\maketitle
\vspace{-3.5em} 
\begin{figure}[H]
\hsize=\textwidth 
\centering
\includegraphics[width=\textwidth]{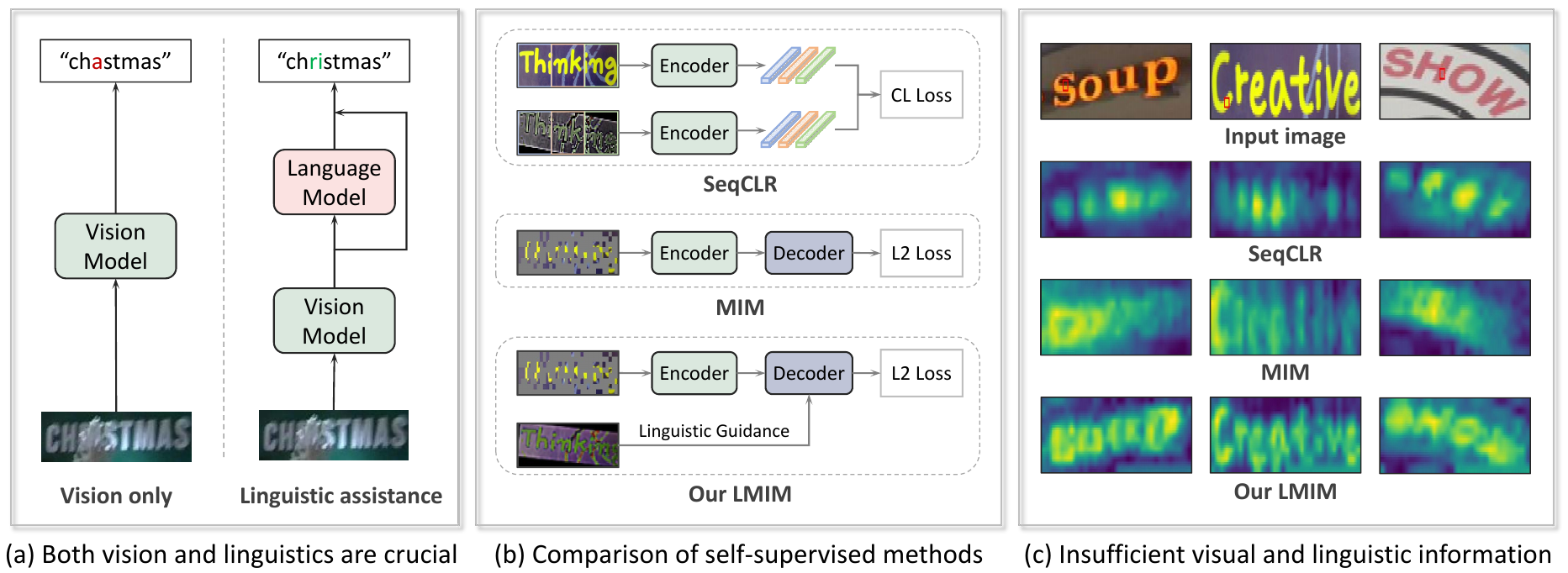}
\caption[]{Illustration of our motivation. (a) Existing studies demonstrate that both visual and linguistic information are crucial for STR, as linguistic information can complement visual features.  (b) Current self-supervised STR approaches, such as sequence contrastive learning (SeqCLR) and masked image modeling (MIM), primarily focus on local region alignment or rely on local visual information for reconstruction, often neglecting the integration of visual and linguistic information at a global level. To address this, our LMIM method channels linguistic information into the decoding process of MIM. (c) Attention maps reveal that SeqCLR lacks character structure information, while MIM emphasizes local regions. Our LMIM effectively captures the global context based on vision and linguistics. The red box in the input image indicates the query.}
\label{fig:motivation}
\end{figure}
}]

\begin{abstract}

Text images are unique in their dual nature, encompassing both visual and linguistic information. The visual component encompasses structural and appearance-based features, while the linguistic dimension incorporates contextual and semantic elements. In scenarios with degraded visual quality, linguistic patterns serve as crucial supplements for comprehension, highlighting the necessity of integrating both aspects for robust scene text recognition (STR). Contemporary STR approaches often use language models or semantic reasoning modules to capture linguistic features, typically requiring large-scale annotated datasets. Self-supervised learning, which lacks annotations, presents challenges in disentangling linguistic features related to the global context. Typically, sequence contrastive learning emphasizes the alignment of local features, while masked image modeling (MIM) tends to exploit local structures to reconstruct visual patterns, resulting in limited linguistic knowledge. In this paper, we propose a Linguistics-aware Masked Image Modeling (LMIM) approach, which channels the linguistic information into the decoding process of MIM through a separate branch. Specifically, we design a linguistics alignment module to extract vision-independent features as linguistic guidance using inputs with different visual appearances. As features extend beyond mere visual structures, LMIM must consider the global context to achieve reconstruction. Extensive experiments on various benchmarks quantitatively demonstrate our state-of-the-art performance, and attention visualizations qualitatively show the simultaneous capture of both visual and linguistic information. The code is available at \url{https://github.com/zhangyifei01/LMIM}. \blfootnote{\Letter ~Corresponding author}
\end{abstract}   
\section{Introduction}
\label{sec:intro}

Visual text plays a crucial role in various real-world applications and encompasses four key areas: text processing~\cite{shuyan_survey_arxiv24,zengweichao_textctrl_nips24,lizhenhang_ecai24}, text detection~\cite{chenyudi_icpr20,qinxugong_mask_mm21,qinxugong_sir_mm23,shuyan_perceiving_mm23}, text recognition~\cite{crnn_tpami17,wangwei_tspnet_mm22,weijin_textblock_mm22,lvjiahao_aaai25}, and text understanding~\cite{zenggangyan_bov_pr23,shenhuawen_divide_ijcai23,shenhuawen_aaai25,zhangyan_aaai25}.
Among these, scene text recognition focuses on extracting textual content from detected text regions in images. Text images uniquely incorporate both visual and linguistic information: visual elements comprise structural and appearance-based features, while linguistic patterns encompass contextual and semantic components~\cite{seed_cvpr20,srn_cvpr20}.  The accuracy of text recognition is challenged by diverse backgrounds, varying lighting conditions, and multiple font styles. The integration of linguistic and visual information is essential for effective recognition, as demonstrated in~\cref{fig:motivation} (a).
Recent approaches have enhanced recognition accuracy by implementing language models~\cite{seed_cvpr20,abinet_cvpr21,matrn_eccv22} or semantic reasoning modules~\cite{srn_cvpr20,visionlan_iccv21,parseq_eccv22}. However, these methods typically require extensive labeled datasets, which are costly and often impractical due to specialized linguistic annotation requirements. Although synthetic data~\cite{synth90k_mj_ijcv16,synthtext_st_cvpr16} partially addresses this issue, a performance gap remains between models trained on synthetic versus real data~\cite{str_fewlabel_cvpr21,union14m_sstr_iccv23}. 
To bridge this gap, leveraging large-scale unannotated real-world data is of practical significance.

Self-supervised learning enables models to learn rich feature representations from real-world scenes~\cite{simclr_icml20,moco_cvpr20,vcp_aaai20,prp_cvpr20,smp_tnnls23,fangbo_mamico_mm22}.
Although specialized self-supervised approaches for STR have emerged, they exhibit limitations in processing both visual and linguistic information at the global level, as illustrated in~\cref{fig:motivation} (b) and (c).
For example, sequence-to-sequence contrastive learning~\cite{seqclr_cvpr21} and character-to-character distillation~\cite{ccd_sstr_iccv23} focus solely on aligning local features at the fragment or character  level without understanding the overall linguistic structure of the text.  
Meanwhile, MIM approaches, such as DiG~\cite{dig_sstr_mm22} and MAERec~\cite{union14m_sstr_iccv23}, prioritize visual pattern reconstruction over coherent text, primarily utilizing local visual features while disregarding global context of characters within the same word.
These limitations highlight the need for better integration of global linguistic information in self-supervised learning for STR, as seen in~\cref{fig:motivation}(c).

In this paper, we propose a simple yet effective Linguistics-aware Masked Image Modeling (LMIM) approach that directly addresses these challenges by concurrently capturing visual structure and linguistic information.
While MIM traditionally reconstructs using local visual features, it inherently models intra-character and inter-character associations, contributing to visual structure comprehension. 
Our LMIM incorporates linguistic information into the decoding process of MIM via a separate branch. 
Specifically, we design a linguistics alignment module that utilizes another image with identical text content but different visual appearances as input to the encoder, alongside the masked image, to extract vision-independent features as linguistic guidance. 
Due to the significant differences in visual appearance, the extracted features are no longer purely visual, requiring LMIM to utilize global context information for reconstruction.
In this way, our method not only maintains the visual structure modeling capability of MIM through reconstruction, but also improves the global context awareness capability by leveraging linguistic information.

Extensive experiments demonstrate our method's effectiveness, achieving state-of-the-art results on both English and Chinese benchmarks. Specifically, LMIM achieves 86.3\% average accuracy on the Union14M benchmark and 97.0\% on six common benchmarks using the ViT-S architecture. Additionally, to address the scarcity of public Chinese text datasets for pre-training, we collected 11 million cropped text images from the Web. Our method achieves impressive results on the Chinese benchmark as well, exhibiting a notable performance of 83.6\% on Scene dataset and 82.0\% on Web dataset. 
These results confirm that by explicitly incorporating linguistic information, LMIM ensures more accurate scene text recognition.

The contributions are summarized as follows: 
\begin{itemize}
    \item We propose a novel method called LMIM to integrate linguistic information into visual information modeling in a self-supervised manner, considering that both types of information are crucial for STR.
    \item Through the designed linguistics alignment module, we extract vision-independent features, compelling LMIM to exploit the global context for reconstruction.
    \item Given the widespread use of Chinese recognition but the lack of large-scale public datasets for pre-training, we contribute a dataset containing 11 million unlabeled Chinese text images for self-supervised STR research.    
    \item Extensive experimental results demonstrate that LMIM achieves state-of-the-art performance on both English and Chinese benchmarks.
\end{itemize}

\section{Related Works}
\label{sec:relatedwork}

\subsection{Scene Text Recognition}

STR presents distinct challenges due to its inherent duality of visual and linguistic information. Existing methods can be categorized into language-free and language-based approaches based on their utilization of language knowledge.
Language-free methods rely exclusively on visual features for prediction. For instance, CRNN~\cite{crnn_tpami17} and TRBA~\cite{strbenchmark_iccv19} employ CNNs for visual feature extraction and RNNs or attention mechanisms for sequence modeling. 
ViTSTR~\cite{vitstr_icdar21} proposes a straightforward framework that utilizes vision transformers.  
In addition, segmentation-based methods~\cite{texstscanner_aaai20,siga_cvpr23} implement fully convolutional networks or specialized architectures for character-level segmentation.
Language-based methods leverage external or internal-learned language representations to assist in text recognition.
The semantic reasoning network~\cite{srn_cvpr20} incorporates a global semantic reasoning module to capture contextual information.
Semantics enhanced encoder-decoder framework~\cite{seed_cvpr20} explicitly integrates pre-trained language models to achieve semantic understanding. 
Subsequently, autonomous, bidirectional and iterative language modeling~\cite{abinet_cvpr21} implements iterative refinement for enhanced semantic reasoning. Visual language modeling network~\cite{visionlan_iccv21} unifies visual and linguistic capabilities within a single vision model using weakly supervised masking.
Following these developments, numerous approaches have explored novel strategies for linguistic information integration~\cite{pimnet_mm21,conclr_aaai22,gtr_aaai22,matrn_eccv22,parseq_eccv22,svtr_ijcai22,clipocr_mm23}.

While STR has achieved remarkable success through large-scale supervised training incorporating both visual and linguistic features, recent research has begun exploring self-supervised learning paradigms for more robust pre-trained models~\cite{seqclr_cvpr21,dig_sstr_mm22}. 

\subsection{Self-supervised Learning}

Self-supervised learning leverages the intrinsic data structure to design pretext tasks, which effectively utilizes large-scale unlabeled data to capture generic representations~\cite{contextpred_iccv15,rotnet_iclr18,instance_discrimination_cvpr18,pcp_icpr20,vcp_aaai20,smp_tnnls23}
These representations are important for a wide range of downstream tasks~\cite{moco_cvpr20,fangbo_uatvr_iccv23,eme_liuchang_tnnls24}.
Recent advancements in contrastive learning (CL) and masked image modeling (MIM) have achieved significant success. The emergence of momentum contrast~\cite{moco_cvpr20} and SimCLR~\cite{simclr_icml20} has spurred extensive research on CL~\cite{debiascl_nips20,jigclu_cvpr21,densecl_cvpr21,ressl_nips21,hcsc_cvpr22,reco_tmm24}. With the advent of vision transformer~\cite{vit_iclr21} and the inspiration of masked language modeling~\cite{bert_arxiv18}, concurrent works such as bidirectional encoder representation from image transformers~\cite{beit_iclr22}, masked autoencoders~\cite{mae_cvpr22}, and SimMIM~\cite{simmim_cvpr22} have demonstrated the effectiveness of MIM. Since then, the research on MIM has dominated the self-supervised learning community~\cite{maskfeat_cvpr22,bootmae_eccv22,ijepa_cvpr23,hardpatch_cvpr23,siammae_nips23,dmjd_liuchang_tmm24,cropmae_eccv24}.

To enhance scene text recognition, existing research leverages the unique characteristics of text images to design self-supervised methods.
Specifically, sequence-to-sequence contrastive learning (SeqCLR)~\cite{seqclr_cvpr21} utilizes the sequential structure prior of data to design contrastive loss. Perceiving stroke-semantic context~\cite{persec_sstr_aaai22} improves SeqCLR by learning representation from low-level strokes. 
Character-to-character distillation~\cite{ccd_sstr_iccv23} introduces character-level contrastive learning with a spectral character segmentation module.
Similarity-aware normalization~\cite{siman_sstr_cvpr22} decouples style and content features by reconstructing image patches guided by neighboring patches.  
Text-degradation invariant auto encoder~\cite{textdiae_aaai23} optimizes three pretext tasks (\textit{i.e.}, masking, blur and background noise) simultaneously.
Dual masked autoencoder~\cite{dualmae_sstr_icdar23} decouples visual and semantic features to explicitly capture text semantics. Symmetric superimposition modeling~\cite{ssm_ijcai24} further emphasizes linguistic information by reconstructing direction-specific signals from symmetrically superimposed input. 
Discriminative and generative self-supervised method~\cite{dig_sstr_mm22} combines CL and MIM, inspired by human learning through reading and writing.  
Unlike these methods, our approach seeks linguistic guidance to capture both character structures and inter-character associations.

\section{Methodology}
\label{sec:method}
\begin{figure*}
  \centering
  \includegraphics[width=2\columnwidth]{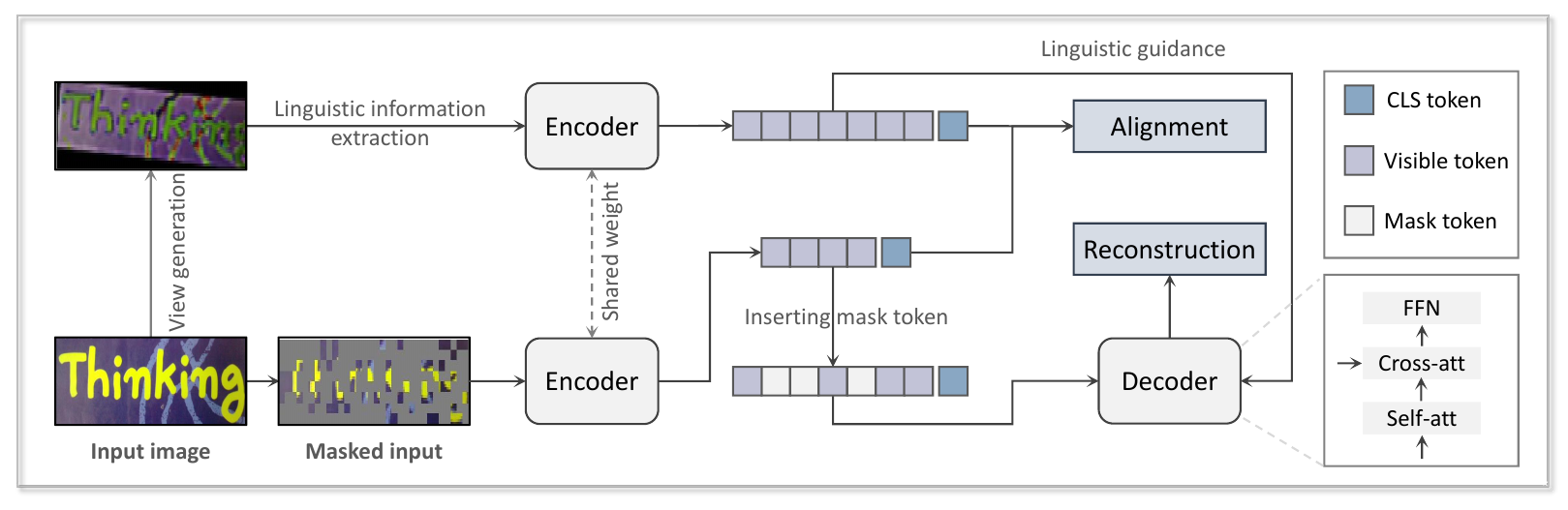}
  
    \caption[]{Overview of our framework. Based on the dual-branch structure, the reconstruction loss and alignment loss are jointly optimized.}
    \label{fig:framework}
\end{figure*}

Although masked image modeling learns both intra-character and inter-character patterns, it primarily focuses on visual structure information and lacks comprehensive linguistic knowledge. Our goal is to enhance the model's ability to understand and integrate richer linguistic information.

\subsection{Overview}

\textbf{ Baseline.} Our method is built upon MAE~\cite{mae_cvpr22}, a seminal work in the field of self-supervised learning, which consists of four core elements: masking strategy, encoder, decoder, and reconstruction target.

Given an image $X \in \mathbb{R}^{H \times W \times C}$, where $H$ and $W$ denote the height and width of the image and $C$ denotes the channel, we split it into $N=H \times W / P^2$ non-overlapping patches $\{x^1, x^2, \cdots, x^N\}$ with the patch size of $P \times P$.
Let $\mathcal{M}$ be the index set of masked patches, $X_v = \{x^k|k \notin \mathcal{M}\}$ denotes the set of visible patches and $X_m = \{x^k|k \in \mathcal{M} \}$ denotes the set of masked patches. The default mask ratio for MAE is 75\%. Only unmasked patches are fed into the ViT encoder and the resulting feature is  $\mathbb{E}(\{x^{cls}, X_v\})$, where $\mathbb{E}$ is the encoder. The features fed into the decoder are the tokens of the visible patches obtained by the encoder and the mask tokens corresponding to the mask patches. The mask token is placed in the corresponding mask position. The output of the decoder is $\{p^{cls}, p^1, p^2, \cdots, p^N\}$.
Mean squared error (MSE) loss is applied to the masked tokens to compute the loss with respect to the corresponding original pixels.
The objective is defined as
\begin{equation}
  \mathcal{L}_{recon} = \frac{1}{||\mathcal{M}||}\sum_{k \in \mathcal{M}} || p^k - t^k || _2 ^2,
  \label{eq:mae}
\end{equation}
where $t^k$ is the reconstruction target, MAE uses original pixels (\textit{i.e.}, $x^k$) by default.

\textbf{Pipeline.}
As shown in~\cref{fig:framework}, our method consists of two branches, \textit{i.e.}, the masked reconstruction branch and the linguistic guidance branch. The input image of the linguistic guidance branch is $\hat{X}$ that contains the same linguistic information as $X$ but different visual appearance. The two encoders share parameters. 
The linguistic information channels the MIM reconstruction via a specially designed decoder. We design a linguistics alignment module to extract vision-independent features as linguistic guidance.
The overall loss function of LMIM is a combination of the reconstruction loss $\mathcal{L}_{recon}$ and the alignment loss $\mathcal{L}_{align}$, which can be formulated as
\begin{equation}
  \mathcal{L} = \mathcal{L}_{recon} +\mathcal{L}_{align} .
  \label{eq:lim}
\end{equation}
The reconstruction target selects the features obtained by the encoder of MAE.

\subsection{Linguistics-aware Masked Image Modeling}
The innovation of our approach lies in integrating linguistic cues into visual information modeling, aiming to enhance the global context understanding of images. 
Our method introduces a novel approach that leverages visual and linguistic information concurrently, comprising two branches: the masked reconstruction branch and the linguistic guidance branch.

\textbf{Guidance View Generation.}
In the self-supervised setting, where additional information is limited, data augmentation becomes crucial. By generating a guidance view that retains comprehensive linguistic information, we address the challenge of integrating linguistic cues within visual data. Previous research highlights the significance of data augmentation in contrastive learning for robust representations~\cite{simclr_icml20}. Unlike traditional methods, which risk misalignments with overly strong augmentations, our approach utilizes varied visual representations of identical textual content, ensuring linguistic coherence without explicit sequence contrastive learning.

\textbf{Linguistics Alignment.}
To decouple vision-independent features, we exploit inputs with different visual representations to learn linguistics consistency.
Specifically, we design the [\textit{cls}] token to encapsulate linguistic information effectively. An alignment loss is introduced to align the [\textit{cls}] token feature:
\begin{equation}
  \mathcal{L}_{align} =  || f^{cls} - \hat{f}^{cls} || _2 ^2.
  \label{eq:limalign}
\end{equation}
This consistency ensures that linguistic information is consistently integrated across different branches, thus forcing the model to learn global linguistic information in addition to visual structures.

\textbf{Linguistics-guided Reconstruction.}
We address the balance between mask ratio and reconstruction complexity. A high mask ratio can complicate reconstruction, while a low ratio might lead to reliance on visual shortcuts. Our default random masking strategy maximizes masking efficiency. Although block masking is feasible, it requires lower ratios and involves higher computational costs. 

The linguistics-guided branch processes unmasked images, ensuring the integrity of linguistic information.
The features obtained from the linguistic guidance branch are denoted as $\mathbb{E}(\{\hat{x}^{cls}, \hat{X}\}) = \{\hat{f}^{cls}, \hat{f}^1, \cdots, \hat{f}^N \}$, with $\hat{X}$ as the input image. For simplicity, we denote $\mathbb{E}(\{\hat{x}^{cls}, \hat{X}\})$ as $\hat{F}$. The sequence of features from the masked reconstruction branch, post-insertion of mask tokens, is recorded as $F$. To integrate linguistic information from $\hat{F}$ into $F$, we design self-attention and cross-attention architecture within the decoder. The attention calculation is represented as:
\begin{equation}
  \mathbb{A} (Q_F, K_F, V_F) =  Softmax(Q_F \cdot K_F^{T}/ \sqrt{d}) \cdot V_F,
  \label{eq:att}
\end{equation}
where $Q_F$, $K_F$, and $V_F$ are derived from $F$ via distinct linear layers, with $d$ as the dimension of $F$. The decoder output is $\mathbb{A} (Q_{\mathbb{A} (Q_F, K_F, V_F)}, K_{\hat{F}}, V_{\hat{F}})$.
The final output is used to calculate the reconstruction loss $\mathcal{L}_{recon}$.

\subsection{Discussion}

\textbf{Innovations and Effectiveness.}
STR necessitates the integration of both visual and linguistic information. 
In self-supervised settings, linguistic information primarily refers to character correlations.
Existing MIM can capture both intra-character and inter-character relationships, but tends to exploit local visual structures to complete reconstruction, resulting in limited global linguistics understanding.
Our LMIM introduces linguistic information into the decoding process of MIM, thus effectively combining visual and linguistic information.
To prevent MIM from exploiting visual structures for reconstruction, we design a linguistic alignment module to disentangle visual-independent features. 
This approach compels MIM to learn global context information rather than relying solely on local visual features for reconstruction. These elements collectively enable the model to effectively integrate linguistic information into visual modeling, achieving enhanced global context perception. Visualization of attention maps clearly demonstrates the advantage of our approach in simultaneously modeling visual and linguistic information, as shown in~\cref{fig:motivation} (c).

\textbf{Limitations.}
The current random masking strategy is suboptimal. Given the variable number of characters in text images, a fixed patch size masking strategy may be inadequate. Future work will explore character density-based masking strategies. Additionally, the pre-training for mask image modeling in STR is restricted to transformer architectures and is inapplicable to CNN architectures.

\section{Experiments}
\label{sec:experiment}

\subsection{Datasets}

\textbf{Pre-training Data.}
We conduct pre-training on real English text data (Union14M-U), and real Chinese text data (our collection).
Union14M-U, a subset of Union14M~\cite{union14m_sstr_iccv23}, includes 10 million unlabeled real images.
To verify the effectiveness on Chinese data, we collected 5 million images from the web and processed them to obtain 11 million unlabeled cropped Chinese text images, named Unlabeled Chinese Text Image 11M (UCTI-11M), for pre-training.

\textbf{Fine-tuning Data.}
Annotated real data (ARD)~\cite{dig_sstr_mm22} contains about 2.8 million annotated real images from TextOCR~\cite{textocr_cvpr21} and Open Image v5.
Union14M-L, another subset of Union14M~\cite{union14m_sstr_iccv23}, comprises approximately 3.2 million labeled real images for fine-tuning.
The Chinese benchmark~\cite{chinese_benchmark_arxiv21} employs about 1.1 million images across four categories (\textit{i.e.}, Scene, Web, Document, and Handwriting) for fine-tuning. 
\setlength{\tabcolsep}{2.3pt}
\begin{table}
  \centering
  \begin{tabular}{@{}ll|ccccccc|c@{}}
    \toprule
    Guide &Align  &Cur. &M-O &Art. &Ctl. &Sal. &M-W &Gen. &Avg \\
    \midrule
    - &- &79.5 &70.2 &71.4 &80.7 &78.6 &82.9 &80.7 &77.7 \\
    $\checkmark$ &- &84.5 &77.1 &73.9 &82.4 &80.6 &83.6 &81.9 &80.6 \\
    $\checkmark$ &$\checkmark$  &85.0 &77.2 &74.6 &83.3 &82.2 &84.0 &81.9 &\textbf{81.2}\\
        \bottomrule
  \end{tabular}
  \caption{Effect of linguistic guidance branch and the alignment loss.}
  \label{tab:componet}
\end{table}
\setlength{\tabcolsep}{3pt}
\begin{table}
  \centering
  \begin{tabular}{@{}l|ccccccc|c@{}}
    \toprule
    Augment  &Cur. &M-O &Art. &Ctl. &Sal. &M-W &Gen. &Avg \\
    \midrule
    Strong &82.9 &74.2 &71.8 &81.0 &79.8 &83.6 &81.5 &79.3 \\
    Weak &82.9 &74.9 &71.1 &80.9 &80.0 &83.1 &81.3 &79.2 \\
    Medium &85.0 &77.2 &74.6 &83.3 &82.2 &84.0 &81.9 &\textbf{81.2}\\
    
        \bottomrule
  \end{tabular}
  \caption{Comparison of different levels of data augmentation.}
  \label{tab:augmentation}
\end{table}
\setlength{\tabcolsep}{2.3pt}
\begin{table}
  \centering
  \begin{tabular}{@{}l|ccccccc|c@{}}
    \toprule
    Decoder  &Cur. &M-O &Art. &Ctl. &Sal. &M-W &Gen. &Avg \\
    \midrule
    CA-SA-FFN &82.6 &74.6 &73.3 &83.2 &81.7 &84.0 &81.8 &80.2 \\
    SA-CA-FFN &85.0 &77.2 &74.6 &83.3 &82.2 &84.0 &81.9 &\textbf{81.2} \\
        \bottomrule
  \end{tabular}
  \caption{Comparison of different decoder architecture designs.}
  \label{tab:decoderarch}
\end{table}

\textbf{Benchmarks.}
The six commonly used benchmarks include three regular text datasets (\textit{i.e.}, IIIT5K-Words (IIIT5K)~\cite{iiit_5k_bmvc12}, ICDAR2013 (IC13)~\cite{ic13_icdar13}, and  Street View Text (SVT)~\cite{svt_iccv11}) and three irregular text datasets (\textit{i.e.}, ICDAR2015 (IC15)~\cite{icdar15_icdar15}, SVT Perspective (SVTP)~\cite{svtp_iccv13}, CUTE80 (CUTE)~\cite{cute80_eswa14}). Recent research identified mislabeled images and performed further verification~\cite{union14m_sstr_iccv23,xiaomeng_spl24}.
The Union14M benchmark contains approximately 0.41 million images with seven challenging scenarios: Curve, Multi-oriented, Artistic, Contextless, Salient, Multi-words, and General text.
The Chinese benchmark~\cite{chinese_benchmark_arxiv21} consists of 0.15 million labeled images across 4 categories.

\subsection{Implementation Details}

\textbf{Pre-training Phase.}
The pre-training phase employs an AdamW optimizer with a cosine learning rate scheduler, utilizing a learning rate of 3e-4 and a batch size of 512. The architecture incorporates a ViT-small encoder~\cite{vit_iclr21} comprising 12 transformer blocks as the default configuration. The decoder uses two layers of blocks composed of self-attention and cross-attention. Input images are processed at a resolution of 32 $\times$ 128 pixels. The masking strategy implements random masking of 80\% of patches with each patch size of 4 $\times$ 4. The reconstruction target utilizes features obtained by the MAERec~\cite{union14m_sstr_iccv23} encoder with a dimension of 384. Unless otherwise specified, the model is trained for 10 epochs by default, including 1 warm-up epoch.

\textbf{Fine-tuning Phase.}
Our baseline text recognition model comprises a ViT encoder and a transformer decoder. Following DiG~\cite{dig_sstr_mm22}, the decoder includes 6 transformer blocks with an embedding dimension of 512. The input image size remains 32 $\times$ 128. We apply the same data augmentation as ABINet~\cite{abinet_cvpr21} during training. 
We use the AdamW optimizer with a learning rate of 2e-4 and a batch size of 512. English data is fine-tuned for 10 epochs and Chinese data for 60 epochs. The maximum sequence length is 25 for English and 40 for Chinese.

\setlength{\tabcolsep}{2.8pt}
\begin{table}
  \centering
  \begin{tabular}{@{}l|ccccccc|c@{}}
    \toprule
    Target  &Cur. &M-O &Art. &Ctl. &Sal. &M-W &Gen. &Avg \\
    \midrule
    Pixel &79.6 &70.1 &70.6 &78.8 &77.0 &78.8 &80.0 &76.4\\
    Rand Feat &72.1 &61.9 &64.6 &76.5 &71.2 &75.9 &78.2 &71.5\\
    MAE Feat &85.0 &77.2 &74.6 &83.3 &82.2 &84.0 &81.9 &\textbf{81.2} \\
        \bottomrule
  \end{tabular}
  \caption{Comparison of different reconstruction targets.}
  \label{tab:target}
\end{table}
\setlength{\tabcolsep}{2pt}
\begin{table}
  \centering
  \begin{tabular}{@{}ll|ccccccc|c@{}}
    \toprule
    Mask &Ratio &Cur. &M-O &Art. &Ctl. &Sal. &M-W &Gen. &Avg \\
    \midrule
    Rand &0.6 &82.9 &74.9 &69.7 &82.2 &79.9 &82.9 &81.5 &79.1 \\
    Rand &0.8 &85.0 &77.2 &74.6 &83.3 &82.2 &84.0 &81.9 &\textbf{81.2} \\
    Rand &0.9 &84.3 &75.0 &74.8 &82.3 &82.0 &82.6 &82.0 &80.4 \\
    Block &0.6 &84.8 &76.8 &75.7 &83.6 &80.6 &83.7 &82.0 &81.0 \\
    Block &0.8 &82.3 &73.6 &74.8 &83.6 &81.0 &83.4 &81.7 &80.1 \\
        \bottomrule
  \end{tabular}
  \caption{Comparison of different masking strategies and ratios.}
  \label{tab:mask}
\end{table}

\textbf{Evaluation Metrics.}
For English text, we evaluate the text recognizer using the WAICS (Word Accuracy Ignoring Case and Symbols) metric by default, which ignores symbols and is case-insensitive.
For Chinese text, we follow existing benchmark settings to calculate sentence-level accuracy for each subset~\cite{chinese_benchmark_arxiv21}.
Average accuracy is calculated using the arithmetic mean (Avg) for subsets and the weighted average (W-Avg) for all instances.

\setlength{\tabcolsep}{4.6pt}
\begin{table*}
  \centering
  \begin{tabular}{@{}llll|cccc|ll|c@{}}
    \toprule
    Method &Publisher &Source &Pre-train &Scene &Web &Document &Handwriting &Avg &W-Avg &Params\\
    \midrule
    MASTER~\cite{master_pr21} &\textit{PR} 21 &\cite{chinese_benchmark_arxiv21} &No &62.1 &53.4 &82.7 &18.5 &54.2$^*$ &61.4$^*$ &63M \\
    ABINet~\cite{abinet_cvpr21} &\textit{CVPR} 21 &\cite{chinese_benchmark_arxiv21} &No &60.9 &51.1 &91.7 &13.8 &54.4$^*$ &62.9$^*$ &53M \\
    TransOCR~\cite{transocr_cvpr21} &\textit{CVPR} 21 &\cite{chinese_benchmark_arxiv21} &No &67.8 &62.7 &97.9 &51.7 &70.0$^*$ &74.8$^*$ &84M \\
    SVTR-B~\cite{svtr_ijcai22} &\textit{IJCAI} 22 &\cite{dctc_aaai24} &No &71.4 &64.1 &99.3 &50.0 &71.2$^*$ &76.6$^*$ &25M \\
    SVTR-L~\cite{svtr_ijcai22} &\textit{IJCAI} 22 &\cite{dctc_aaai24} &No &72.1 &66.3 &99.3 &50.3 &72.0$^*$ &77.2$^*$ &41M \\
    CIRN~\cite{cirn_ctr_ijcai23} &\textit{IJCAI} 23 &\cite{cirn_ctr_ijcai23} &No &73.3 &- &- &- &- &- &- \\
    DCTC-B~\cite{dctc_aaai24} &\textit{AAAI} 24 &\cite{dctc_aaai24} &No &72.2 &67.0 &\textbf{99.4} &50.4 &72.3$^*$ &77.3$^*$ &25M \\
    DCTC-L~\cite{dctc_aaai24} &\textit{AAAI} 24 &\cite{dctc_aaai24} &No &73.9 &68.5 &\textbf{99.4} &51.0 &73.2$^*$ &78.3$^*$ &41M \\
    LMIM (Scratch) &\textit{CVPR} 25 &Ours &No  &75.1 &74.7 &97.1 &53.3 &75.1 &79.0  &36M \\
    \midrule
    TransOCR~\cite{transocr_cvpr21} &\textit{CVPR21} &\cite{chinese_benchmark_arxiv21} &Yes &68.5 &62.5 &97.9 &53.5 &70.6$^*$ &75.3$^*$ &84M \\
    CCR-CLIP~\cite{ccr_clip_iccv23} &\textit{ICCV} 23 &\cite{ccr_clip_iccv23} &Yes &71.3 &69.2 &98.3 &60.3 &74.8$^*$ &78.3$^*$ &-  \\
    MaskOCR-S~\cite{maskocr_tmlr24} &\textit{TMLR} 24 &\cite{maskocr_tmlr24} &Yes &71.4 &72.5 &98.8 &55.6 &74.6$^*$ &78.1 &36M \\
    MaskOCR-B~\cite{maskocr_tmlr24} &\textit{TMLR} 24 &\cite{maskocr_tmlr24} &Yes &73.9 &74.8 &99.3 &63.7 &77.9$^*$ &80.8  &100M \\
    \midrule
    SeqCLR~\cite{seqclr_cvpr21} &\textit{CVPR} 21 &Ours & Yes &81.7 &80.5 &98.5 &60.3 &80.3 &83.8 &36M \\
    MIM~\cite{mae_cvpr22} &\textit{CVPR} 22 &Ours & Yes &82.3 &80.9 &98.9 &62.4 &81.1 &84.6 &36M  \\
    LMIM &\textit{CVPR} 25 &Ours &Yes &\textbf{83.6} &\textbf{82.0} &99.1 &\textbf{63.9} &\textbf{82.2} &\textbf{85.5} &36M  \\
    \bottomrule
  \end{tabular}
  \caption{Results on Chinese benchmark. $^*$ means results re-calculated using those works' original results by us.}
  \label{tab:chinesebench}
\end{table*}

\setlength{\tabcolsep}{1pt}
\begin{table*}
  \centering
  \begin{tabular}{@{}llll|ccccccc|cc|c@{}}
    \toprule
    Method &Publisher &Source &PT-data &Curve &M-O &Artistic &Contextless &Salient &M-W &General &Avg &W-Avg &Params\\
    \midrule
    ABINet~\cite{abinet_cvpr21} &\textit{CVPR} 21 &\cite{union14m_sstr_iccv23} &- &75.0 &61.5 &65.3 &71.1 &72.9 &59.1 &79.4 &69.2 &79.2$^*$ &37M \\
    VisionLAN~\cite{visionlan_iccv21} &\textit{ICCV} 21 &\cite{union14m_sstr_iccv23} &- &70.7 &57.2 &56.7 &63.8 &67.6 &47.3 &74.2 &62.5 &74.0$^*$ &33M \\
    MATRN~\cite{matrn_eccv22} &\textit{ECCV} 22 &\cite{union14m_sstr_iccv23} &- &80.5 &64.7 &71.1 &74.8 &79.4 &67.6 &77.9 &74.6 &77.8$^*$ &44M \\
    PARSeq~\cite{parseq_eccv22} &\textit{ECCV} 22 &\cite{ssm_ijcai24} &- &79.8 &79.2 &67.4 &77.4 &77.0 &76.9 &80.6 &76.9 &80.5$^*$ &24M \\
    DiG-S~\cite{dig_sstr_mm22} &\textit{ACMMM} 22 &\cite{ssm_ijcai24} &U14M-U &85.9 &83.5 &77.4 &82.5 &84.3 &84.0 &83.8 &83.0 &83.8$^*$ & 36M \\
    MAERec-S~\cite{union14m_sstr_iccv23} &\textit{ICCV} 23 &\cite{union14m_sstr_iccv23} &U14M-U &81.4 &71.4 &72.0 &82.0 &78.5 &82.4 &82.5 &78.6 &82.4$^*$ &36M \\
    
    SSM-S~\cite{ssm_ijcai24} &\textit{IJCAI} 24 &\cite{ssm_ijcai24} &U14M-U &87.5 &85.8 &78.4 &84.8 &85.2 &85.0 &84.0 &84.3 &84.0$^*$ &36M \\
    \midrule
    SeqCLR~\cite{seqclr_cvpr21} &\textit{CVPR} 21 &Ours & U14M-U &83.7 &79.9 &73.7 &79.7 &81.0 &84.0 &82.7 &80.7 &82.7 &36M \\
    MIM~\cite{mae_cvpr22} &\textit{CVPR} 22 &Ours & U14M-U &84.8 &79.3 &71.1 &81.5 &81.0 &82.5 &82.0 &80.3 &82.0 &36M \\
    LMIM$^\ddag$ &\textit{CVPR} 25 &Ours &U14M-U &89.7 &84.9 &79.4 &81.9 &86.0 &82.9 &85.1 &84.3 &85.1 &36M  \\
    LMIM &\textit{CVPR} 25 &Ours &U14M-U &87.5 &84.5 &79.8 &84.3 &86.6 &86.1 &84.4 &84.7 &84.4 &36M  \\
    
    \midrule
    LMIM$_{20ep}^\ddag$ &\textit{CVPR} 25 &Ours &U14M-U &\textbf{90.6} &\textbf{86.9} &80.0 &82.3 &87.7 &84.4 &\textbf{85.7} &85.4 &\textbf{85.7} &36M  \\
    LMIM$_{20ep}$ &\textit{CVPR} 25 &Ours &U14M-U &90.3 &86.6 &\textbf{80.7} &\textbf{85.5} &\textbf{88.2} &\textbf{87.9} &85.1 &\textbf{86.3} &85.1 &36M  \\
    \bottomrule
  \end{tabular}
  \caption{Results on English Union14M benchmark. Unless otherwise specified, all text recognizers are trained using real data from Union14M-L. PT-data stands for the pre-training dataset. U14M-U denotes Union14M-U. M-O indicates Multi-Oriented, while M-W indicates Multi-Words. LMIM$_{20ep}$ refers to pre-training for 20 epochs. The symbol $^\ddag$ represents fine-tuning on ARD.}
  \label{tab:union14m}
\end{table*}

\setlength{\tabcolsep}{1.5pt}
\begin{table*}
  \centering
  \begin{tabular}{@{}lllll|cccccc|ll|c@{}}
    \toprule
    Method &Publisher &Source &PT-data &FT-data & IIIT5K &IC13 & SVT &IC15 &SVTP &CUTE &Avg &W-Avg &Params\\
    \midrule
    ABINet~\cite{abinet_cvpr21} &\textit{CVPR} 21 &\cite{union14m_sstr_iccv23} &- &U14M-L &97.2 &97.2 &95.7 &87.6 &92.1 &94.4 &94.0 &93.9$^*$ &37M \\
    VisionLAN~\cite{visionlan_iccv21} &\textit{ICCV} 21 &\cite{union14m_sstr_iccv23} &- &U14M-L &96.3 &95.1 &91.3 &83.6 &85.4 &92.4 &91.3 &91.2$^*$ & 33M \\
    MATRN~\cite{matrn_eccv22} &\textit{ECCV} 22 &\cite{union14m_sstr_iccv23} &- &U14M-L &98.2 &97.9 &96.9 &88.2 &94.1 &97.9 &95.5 &95.0$^*$ &44M \\
    PARSeq~\cite{parseq_eccv22} &\textit{ECCV} 22 &\cite{ssm_ijcai24} &- &U14M-L &98.0 &96.8 &95.2 &85.2 &90.5 &96.5 &93.8 &93.5$^*$ &24M \\
    MaskOCR-S~\cite{maskocr_tmlr24} &\textit{TMLR} 24 &\cite{maskocr_tmlr24} &STD\&URD &ARD &98.0 &97.8 &96.9 &90.2 &94.9 &96.2 &95.6 &95.4$^*$ &31M \\
    DiG-S~\cite{dig_sstr_mm22} &\textit{ACMMM} 22 &\cite{ssm_ijcai24} &U14M-U &U14M-L &98.7 &97.8 &98.5 &88.9 &92.7 &96.5 &95.5 &95.3$^*$ & 36M \\
    DiG-S~\cite{dig_sstr_mm22} &\textit{ACMMM} 22 &\cite{dig_sstr_mm22} &STD\&URD &ARD &97.7 &97.3 &96.1 &88.6 &91.6 &96.1 &94.6$^*$ &94.5$^*$ &36M \\
    CCD~\cite{ccd_sstr_iccv23} &\textit{ICCV} 23 &\cite{ccd_sstr_iccv23} &STD\&URD &ARD &98.0 &\textbf{98.3} &96.4 &90.3 &92.7 &98.3 &95.6$^*$ &95.4$^*$ &36M \\
    MAERec-S~\cite{union14m_sstr_iccv23} &\textit{ICCV} 23 &\cite{union14m_sstr_iccv23} &U14M-U &U14M-L &98.0 &97.6 &96.8 &87.1 &93.2 &97.9 &95.1 &94.5$^*$ &36M \\
    SSM-S~\cite{ssm_ijcai24} &\textit{IJCAI} 24 &\cite{ssm_ijcai24} &U14M-U &U14M-L & 99.0 & \textbf{98.3} &97.8 &89.5 &94.0 &98.3 &96.1 &95.8$^*$ &36M\\
    \midrule
     LMIM &\textit{CVPR} 25 &Ours &U14M-U &ARD &98.0 &97.9 &96.9 &88.2 &93.5 &98.3 &95.5 &94.9 &36M \\
     LMIM &\textit{CVPR} 25 &Ours &U14M-U &U14M-L &98.5 &98.0 &97.7 &88.7 &94.0 &98.6 &95.9 &95.3 &36M \\
    SeqCLR$^\dag$~\cite{seqclr_cvpr21} &\textit{CVPR} 21 &Ours & U14M-U &U14M-L &98.8 &97.9 &96.8 &91.4 &92.9 &96.9 &95.8 &95.9 &36M \\
    MIM$^\dag$~\cite{mae_cvpr22} &\textit{CVPR} 22 &Ours & U14M-U &U14M-L &98.5 &97.6 &96.0 &89.5 &91.3 &96.9 &95.0 &95.1 &36M \\
   
    LMIM$^\dag$ &\textit{CVPR} 25 &Ours &U14M-U &ARD &98.8 &97.8 &97.1 &90.8 &94.4 &97.9 &96.1 &96.0 &36M\\
    LMIM$^\dag$ &\textit{CVPR} 25 &Ours &U14M-U &U14M-L &\textbf{99.3} &98.1 &\textbf{98.3} &\textbf{91.7} &\textbf{95.4} &\textbf{99.3} &\textbf{97.0} &\textbf{96.7} &36M \\
    \bottomrule
  \end{tabular}
  \caption{Results on six common benchmarks. PT-data refers to the pre-training dataset, and FT-data refers to the fine-tuning dataset. U14M-L and U14M-U are Union14M-L and Union14M-U, respectively. URD denotes the unlabeled real dataset containing 15.8 million images. STD refers to 17 million synthetic data. The symbol $^\dag$ denotes using corrected label.}
  \label{tab:sixbench}
\end{table*}

\subsection{Ablation Study}
To efficiently verify the effectiveness of our method, we use two subsets of Union14-U, \textit{i.e.}, Book32 and OpenImages, comprising approximately 5 million images, for both pre-training and fine-tuning over 5 epochs each. 
The seven datasets of Union14M benchmark are abbreviated as Cur., M-O, Art., Ctl., Sal., M-W, and Gen. in the table.

\textbf{Linguistic Guidance and Alignment Loss.}
Our method consists of two parts: reconstruction loss and alignment loss,~\cref{eq:lim}. The effects of these components are shown in~\cref{tab:componet}. Neither loss pertains to single-branch masked reconstruction. Optimal performance is achieved when both losses are optimized simultaneously, resulting in an average accuracy increase of 3.5\% compared to when neither loss is applied. Excluding the alignment loss results in an average accuracy decrease of 0.6\%. These findings underscore the importance of both components.

\textbf{Guidance View Generation.}
We apply varying degrees of data augmentation to the original image to generate the guided view. Weak augmentation involves only geometric transformations, such as cropping and rotation. Medium augmentation adds color transformation, distortion, and perspective transformation to the basic geometric changes. Strong augmentation further adjusts parameters from medium augmentation to enhance the level. Experimental results in~\cref{tab:augmentation} demonstrat the importance of maintaining large visual appearance differences while preserving complete linguistic information.

\textbf{Decoder Architecture.}
In~\cref{tab:decoderarch}, we compare decoder architectural designs. CA-SA-FFN indicates that features from the mask and guide branches are initially processed with cross-attention, followed by self-attention. Conversely, SA-CA-FNN involves self-attention on the mask branch features before interaction with the guide branch features. The experimental results suggest that performing self-attention prior to cross-attention is more effective.


\textbf{Reconstruction Target.}
By default, we use the MAE feature as our reconstruction target. As illustrated in~\cref{tab:target}, we compare different reconstruction targets. 
Increased iterations or specific training strategies can minimize differences between reconstruction target when handling large-scale data~\cite{dbot_iclr24}.

\textbf{Mask Strategy and Ratio.}
Different masking strategies and ratios are crucial to mask reconstruction methods. We compare random and block masking. Unlike random masking, block masking obscures large continuous areas, thereby increasing reconstruction difficulty. As shown in~\cref{tab:mask}, both masking methods are effective with appropriate mask ratios, but block masking requires fewer mask ratios (more patches need to be processed). Therefore, we default to the random masking strategy with a mask ratio of 80\%.


\subsection{Performance Comparison}


\textbf{Chinese Benchmark.}
Compared with other languages, Chinese relies more on linguistic knowledge to understand the content.
To demonstrate the versatility of our method across different languages, we conducted experiments on Chinese data. As shown in~\cref{tab:chinesebench}, our method achieves state-of-the-art performance in terms of average accuracy. In particular, our method achieves a significant performance of 83.6\% on the Scene dataset and 82.0\% on the Web dataset, underscoring its ability to learn linguistic information and its versatility across languages.

\textbf{Union14M Benchmark.}
In~\cref{tab:union14m}, the pre-trained methods (\textit{i.e.}, DiG, MAERec, and SSM) consistently outperform those trained from scratch.
When pre-trained on Union14M-U and subsequently fine-tuned on Union14M-L, our method achieves consistent improvements across all subsets. Notably, we achieve a state-of-the-art average accuracy of 86.3\% when pre-training for 20 epochs, representing a 2.0\% improvement over SSM. 
We also experiment with fine-tuning on the ARD dataset, resulting in an average accuracy of 85.4\%, with the General subset reaching 85.7\%. 
This improvement stems from our method’s design, which integrates visual and linguistic information, enhancing the encoder’s feature representation for STR.

\textbf{Common Benchmarks.}
We evaluate our method using corrected labels across six common benchmarks. Experimental results indicate an average accuracy of 97.0\% and a weighted average accuracy of 96.7\%, achieving new state-of-the-art performance,~\cref{tab:sixbench}.
Notably, higher improvements are observed on three irregular datasets, demonstrating that linguistic information offers additional benefits.

\begin{figure}
  \centering
  \includegraphics[width=\columnwidth]{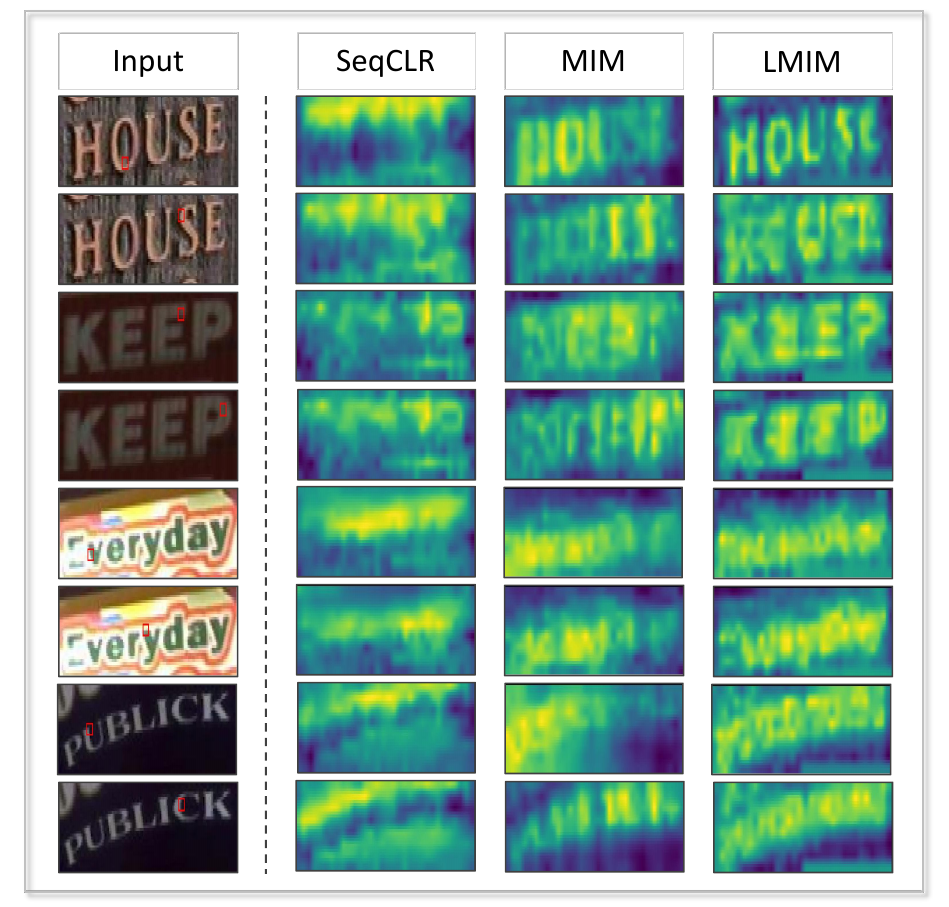}
  
    \caption[]{Visualization of attention maps. The red box in the input image refers to the query.}
    \label{fig:attvis}
\end{figure}

\subsection{Visualization}

To qualitatively verify that our method has effectively learned linguistic information compared to sequence contrastive learning and masked image modeling, we visualize the attention maps for different queries,~\cref{fig:motivation} (c) and~\cref{fig:attvis}. 
We use the attention map obtained from the last block (12-th block) of ViT-Small.
We can see that SeqCLR’s attention at each query position is rather chaotic and cannot reflect the character structure. This is because the contrast of local sequence elements does not fully learn the underlying structure and the association between characters.
In particular, we compare the attention maps obtained for different query positions of the same image in~\cref{fig:attvis}.
The attention maps obtained from different queries of the same image and different queries of different images demonstrate that the MIM method pays more attention to local areas. This is because the visual features of local areas can be used to reconstruct the mask reconstruction process.
Our LMIM not only pays attention to different characters but also shows a clear character structure, proving that the method captures both visual and linguistic information.



\section{Conclusion}
\label{sec:conclusion}

In this paper, we propose Linguistics-aware Masked Image Modeling, \textit{i.e.}, LMIM, a simple yet effective self-supervised learning framework specifically designed for STR. Our approach introduces a dual-branch structure that integrates linguistic cues into visual modeling, considering both simultaneously as crucial for STR. 
A specially designed linguistic alignment module leverages images with varying visual appearances to disentangle visual-independent linguistic features. Unlike methods focused solely on visual structure, LMIM encourages the model to utilize global contextual information for reconstruction. Our method delivers significant improvements on various benchmarks, including English and Chinese text recognition tasks. Attention visualization qualitatively demonstrates that our method effectively combines visual and linguistic information. 
In future work, we will explore more suitable masking strategies tailored to character density, aiming to further optimize model performance and effectively handle diverse scene text recognition tasks.

\section*{Acknowledgement}
\label{sec:acknowledge}

This work is supported by the National Natural Science Foundation of China (Grant NO 62376266 and 62406318), Key Laboratory of Ethnic Language Intelligent Analysis and Security Governance of MOE, Minzu University of China, Beijing, China.


{
    \small
    \bibliographystyle{ieeenat_fullname}
    \bibliography{main}
}


\end{document}